\documentclass{article}

\PassOptionsToPackage{numbers, compress}{natbib}

\usepackage{neurips_2021} %
\usepackage[utf8]{inputenc} %
\usepackage[T1]{fontenc}    %
\usepackage{hyperref}       %
\usepackage{url}            %
\usepackage{booktabs}       %
\usepackage{amsfonts}       %
\usepackage{nicefrac}       %
\usepackage{microtype}      %
\usepackage{amsmath}
\usepackage{amssymb}
\usepackage{amsthm}
\usepackage{amscd}
\usepackage{comment}
\usepackage{graphicx}
\usepackage{multirow}
\usepackage{subcaption}
\usepackage{float}
\usepackage{array}
\usepackage{booktabs} %
\usepackage{xcolor}
\usepackage{color, colortbl}
\usepackage{wrapfig}
\usepackage{enumitem}
\usepackage{tkz-euclide}
\definecolor{Gray}{gray}{0.92}

\newif\ifsubmit
\submitfalse

\ifsubmit
\newcommand{\tianshi}[1]{}
\newcommand{\jingkang}[1]{}
\newcommand{\annie}[1]{}
\newcommand{\siva}[1]{}
\newcommand{\mengye}[1]{}
\else
\newcommand{\tianshi}[1]{{\color{blue}[Tianshi: #1]}}
\newcommand{\jingkang}[1]{{\color{magenta}[Jingkang: #1]}}
\newcommand{\annie}[1]{{\color{orange}[Annie: #1]}}
\newcommand{\siva}[1]{{\color{brown}[Siva: #1]}}
\newcommand{\mengye}[1]{{\color{cyan}[Mengye: #1]}}
\fi
\title{Zero-Shot Compositional Policy Learning via Language Grounding}

\author{\vspace{0.55mm}
Tianshi Cao$^{*1,2}$ \quad Jingkang Wang$^{*1,2}$\quad Yining Zhang$^{1,2}$ \quad \textbf{Sivabalan Manivasagam}$^{1,2}$ \\
{University of Toronto$^{1}$, Vector Institute$^{2}$} \\ 
{\tt\small {\{jcao, wangjk, ynz, manivasagam\}}@cs.toronto.edu}}

\begin{document}
\maketitle
 
\begin{abstract}
Despite recent breakthroughs in reinforcement learning (RL) and imitation learning (IL), existing algorithms fail to generalize beyond the training environments. In reality, humans can adapt to new tasks quickly by leveraging prior knowledge about the world such as language descriptions. 
To facilitate the research on language-guided agents with domain adaption, %
we propose a novel zero-shot compositional policy learning task, where the environments are characterized as a composition of different attributes. 
Since there are no public environments supporting this study, we introduce a new research platform \textit{BabyAI++} in which the dynamics of environments are disentangled from visual appearance. At each episode, \textit{BabyAI++} provides varied vision-dynamics combinations along with corresponding descriptive texts. To evaluate the adaption capability of learned agents, a set of vision-dynamics pairings are held-out for testing on \textit{BabyAI++}.
Unsurprisingly, we find that current language-guided RL/IL techniques overfit to the training environments and suffer from a huge performance drop when facing unseen combinations. In response, we propose a multi-modal fusion method with an attention mechanism to perform visual language-grounding. Extensive experiments show strong evidence that language grounding is able to improve the generalization of agents across environments with varied dynamics. %

\end{abstract}

\section{Introduction}
\vspace{-0.1in}

Reinforcement learning (RL) and imitation learning (IL) have recently witnessed tremendous success in various applications such as game-playing~\cite{mnih2015human,silver2017mastering} and robotics~\cite{DBLP:conf/iros/Chatzilygeroudis17,QuillenJNFIL18,DAGGER}.
However, existing algorithms are commonly trained and evaluated on a fixed environment, impeding generalization to previously unseen environments~\cite{pmlr-v97-cobbe19a}. In contrast, humans adapt to unseen scenarios quickly by leveraging the prior knowledge of the world, such as through language descriptions. For instance, the baby can associate the word \textit{ball} with different images of balls (basketball, baseball, etc.), and then associate common properties with it (round, can bounce, and be thrown). Through language grounding, babies quickly grasp concepts and can generalize learning easily to new scenes. Another notable example is that humans are able to summarize task-relevant knowledge by reading manuals or descriptions about the environments, to learn and generalize across tasks faster (e.g., learning to use new electrical appliances or play new games).
Motivated by human learning, we are interested in building agents that 
generalize beyond training environments by leveraging language descriptions.

Previous works on language-conditioned policy learning have focused on leveraging instruction and goal-based text for environments where the reward function shifts at test time~\cite{LuketinaNFFAGWR19, branavan2012learning, babyai,GoyalNM19}. In such settings, joint representations of visual features and instructive text have been shown to be successful for vision-and-language navigation~\cite{DoomNav,HuFRKDS19,anderson2018vision,Hermann} and question answering~\cite{BAIDUNLPRL}. %
While prior work has focused on using language to help the agent's generalization when the reward function changes at test time, we explore another challenging setting:
using language to inform test-time shifts of the world model dynamics (i.e. novel state transition functions). 
Our setting, while different, is related to past work, as the shifts to model dynamics also affect the reward function. 

To facilitate research on language-guided agents on domains with large test-time shifts, we introduce a novel zero-shot compositional policy learning task:
during training, a language-guided agent learns on an environment with a set of model dynamics that changes from episode to episode. These model dynamics are encoded via visual cues, which we call "vision-dynamics" pairings. Text descriptions about the vision-dynamics pairings are also provided to the agent to help it generalize. 
To evaluate zero-shot learning, at test-time, the language-guided agent will be placed in an environment with the same task, but different vision-dynamics pairings it has not seen before. To perform zero-shot transfer, the agent will need to combine both textual instructions and visual grounding to solve the task.

 Zero-shot compositional learning (ZSCL) has recently been explored in vision as a classification task for concepts combining an object and attribute~\cite{yang2020learning, purushwalkam2019task, atzmon2020causal}, and in language where networks are tasked with executing instructions posed as composed sequences~\cite{lakeNIPS2019}. In RL, ZSCL has been studied in terms of task~\cite{sohn2018hierarchical}, and policy~\cite{sahni2017learning} compositions.
 We extend ZSCL to the RL/IL domains for policy learning, 
 where we compose dynamics and visual attributes through descriptive textual cues.

\begin{wraptable}{R}{0.5\linewidth}
    \centering
    \caption{Comparing our proposed environment with other available environments. 1: The agent may manipulate the environment to achieve its goals; 2: Environment dynamics change per episode; 3: Can generate a wide variety of scenarios for task learning. }
    \vspace{-2mm}
    \label{tab:related}
    \resizebox{0.9\linewidth}{!}{
    \begin{tabular}{c|cccccc}
        Environments & \rotatebox{90}{Instructive Text} & \rotatebox{90}{Descriptive Text} & \rotatebox{90}{Env. Manipulation$^1$} & \rotatebox{90}{Variable Dynamics} & \rotatebox{90}{Procedural Env. $^2$} & \rotatebox{90}{Multi-Task $^3$} \\
        \hline
        AI2-THOR \cite{ai2thor} &  &  & \checkmark & \checkmark & \checkmark &  \\
        ALFRED \cite{ALFRED20} &\checkmark &  & \checkmark & \checkmark &  & \checkmark \\
        House3D \cite{wu2018building} & \checkmark &  &  &  & \checkmark &  \\
        Descriptive GVGAI \cite{TransferRLNLP} & \checkmark & \checkmark & \checkmark & & & \\
        NLP Nav VisDoom \cite{DoomNav} & \checkmark & & & & \checkmark &  \\
        BabyAI \cite{babyai} & \checkmark &  & \checkmark & & \checkmark & \checkmark \\
        \hline
        BabyAI++ (Ours) & \checkmark & \checkmark & \checkmark & \checkmark & \checkmark & \checkmark \\
    \end{tabular}}
    \vspace{-4mm}
\end{wraptable}

Existing RL environments are limiting for this setting. Learning with descriptive text has only been explored in limited scope with hand-engineered language features and few environments~\cite{branavan2012learning, TransferRLNLP}. More extensive language-RL platforms~\cite{ai2thor, wu2018building, babyai} only provide language to the agent via goal-oriented textual instructions. %
To meet the needs for ZSCL with varying model dyamics and text descriptions, we extend the BabyAI \cite{babyai} environment and propose BabyAI++ for ZSCL. 
BabyAI~\cite{babyai} is an open-source grid-world environment that focuses on grounded language learning in RL/IL agents by creating procedurally-generated task descriptions and corresponding environments of varying levels of complexity.
BabyAI++ adds two important components for performing ZSCL which BabyAI~\cite{babyai} currently lacks: changing model dynamics for the same task, and automatically generated text descriptions that define the model dynamics.

We evaluate current language-guided baselines on BabyAI++ and find that existing techniques fail to adapt to composition changes.
As a response, we develop a multi-modal fusion algorithm with an attention mechanism to perform visual language-grounding. Specifically, the language embeddings are assigned to different locations in the scene feature maps. %
Extensive evaluations on BabyAI++ tasks show that the proposed approach 
improves over baselines across a range of tasks,
especially when the complexity of the environments and unseen attributes grows. Interpretable results on the trajectories and attention module provide strong evidence that our method is able to learn compositional concepts in a zero shot manner by grounding language. We hope our platform will help inspire more future research on zero-shot compositional RL and IL problems with language grounding.

\vspace{-0.1in}
\section{Related work}
\vspace{-0.1in}
We study zero-shot compositional learning (ZSCL) in the RL/IL setting via language grounding, which is closely related to language-conditioned RL and ZSCL classification tasks. Moreover, we focus on the generalization issue of RL/IL and show descriptive language is useful.
Please see supplementary for a detailed comparison of BabyAI++ with existing RL environments.

\vspace{-0.1in}
\paragraph{Language-conditioned RL} Our work evolves from current approaches in language-conditioned RL by considering the grounding of environment descriptions prior to instruction following.
Specifically, previous studies in interactive robotics have produced platforms for instruction following \cite{6696569,5650910,10.5555/1734454.1734553} in diverse environments. 
Another popular branch is vision-and-language navigation (VLN), which deals specifically with the task of navigating to language specified locations within the environment \cite{Hermann, mirowski2018learning, Brahmbhatt_2017, anderson2018vision, DoomNav, hermann2017grounded, brodeur2017home, wu2018building, ALFRED20, gordon2018iqa}.
Recent works also explored the use of natural language in RL beyond the scope of navigation. Branavan et al. \cite{branavan2012learning} utilized the language from game manuals to guide a Monte-Carlo-search-based agent to play \textit{Civilization II}. 
However, these works aim at optimizing a policy in a complex environment rather than learn representative embeddings to generalize to unseen environments.
Most relevant to our work, \cite{BAIDUNLPRL,babyai} propose similar grid-world environments for VLN or visual question answering. However, these environments lack task diversity and have limited interactions between agents and games (see Table~\ref{tab:related}).

\vspace{-0.1in}
\paragraph{Zero-shot compositional learning} 
Our work extends the conventional ZSCL setting to policy learning. The goal of ZSCL is to generalize to nouveau concepts with no training examples through the composition of sub-concepts. Many works have studied this problem in the zero-shot classification domain \cite{yang2020learning, purushwalkam2019task, atzmon2020causal}. Some recent works in RL focus on the composition of skills for solving hierarchical RL tasks \cite{sahni2017learning, sohn2018hierarchical}. In contrast, we consider a different kind of zero-shot composition, namely, nouveau compositions of visual appearances and behaviors of objects in the environment (physical dynamics). Note that synergy between language and visual composition is necessary to complete the tasks since the visual-dynamics links are only provided via descriptive texts.

\vspace{-0.1in}
\paragraph{Generalization in policy learning} Learning a policy that the agent can follow and generalize to unseen environments has attracted attention in recent years. Approaches such as multi-task learning \cite{DBLP:conf/nips/TehBCQKHHP17, yang2017multi, wang2018nervenet}, meta-learning \cite{LEO,snail,maml}, learning to plan \cite{srinivas2018universal, nair2018visual,kurutach2018learning}, and zero-shot learning \cite{oh2017zero,perez2018film}, have all tackled this problem in different ways. A common emerging trend %
is that encountering a wide variety of environments during training 
improves
performance on new environments. Unfortunately, most works are evaluated on environments that differ only slightly from those used in training due to the lack of variety in available benchmarks. This necessitates the development of research platform that can effortlessly synthesize new environments to support research in these directions. %

\vspace{-2mm}
\section{Problem formulation}
\label{sec:babyai}
\vspace{-2mm}

We begin by introducing a policy learning framework with language descriptions. %
Then we introduce the concept of compositional zero-shot policy learning.

\begin{wrapfigure}{R}{0.5\textwidth}
\vspace{-4mm}
\begin{minipage}{0.5\textwidth}
\centering
    \includegraphics[width=\textwidth]{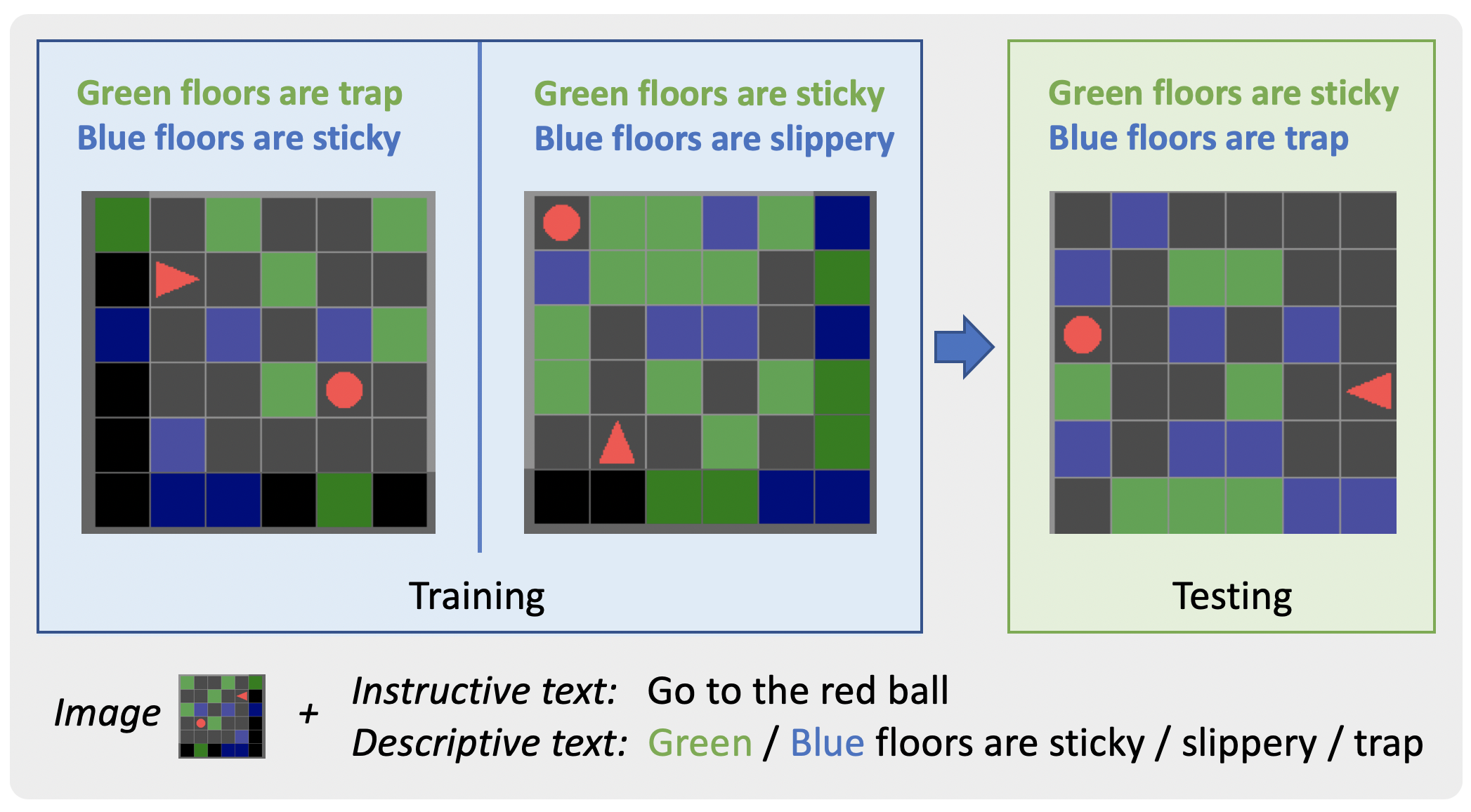}
    \vspace{-5mm}
    \caption{\texttt{GoToRedBall} task in \textit{BabyAI++}.
    }
    \label{fig:babyai++}
\end{minipage}
\vspace{-4mm}
\end{wrapfigure}

\vspace{-0.1in}
\paragraph{Policy learning with language descriptions}
We consider the learning task in a standard policy learning problem with the addition that a description of the model dynamics of the environment is available to the agent. 
An environment is uniquely defined by the environment tuple $\mathbf{E} = \{\mathcal{S}, \mathcal{A}, \mathcal{F}, \rho_0\}$ which consists of the collection of states $\mathcal{S}$, collection of actions $\mathcal{A}$, state transition density function $\mathcal{F}(s,a,s'): \mathcal{S} \times \mathcal{A} \times \mathcal{S} \mapsto \mathbb{R}$, and initial state distribution $\rho_0$. 
Then, a RL task can be defined as an environment equipped with a reward function and related parameters $\mathbf{T} = \{\mathbf{E}, R, \gamma, H\}$, where $R: \mathcal{S} \times \mathcal{A} \mapsto \mathbb{R}$ is the reward function, $\gamma$ is the discount factor, and $H$ is the time horizon of the task. In contrast, an IL task is defined by the addition of an expert policy $\pi_E: \mathcal{S} \times \mathcal{A} \mapsto \mathbb{R}$. The goal of the agent is to mimic this policy.
In this work, we focus on the setting where the environment is augmented with a description of the environment dynamics, $\mathbf{E}_d = \{\mathbf{E}, \mathbf{D}\}$, where $\mathbf{D}$ is a function of $\mathcal{F}$. %
\vspace{-2mm}
\paragraph{Compositional zero-shot policy learning of environment dynamics}
In this work, we 
explore
learning composition of concepts in environment descriptions. 
Below we define the ZSCL setting, and in Sec. \ref{sec:babyai++} we provide a real example of ZSCL (see Figure~\ref{fig:babyai++}).
Consider the case where the environment description expresses environment dynamics as a set of $n$ tuples $\mathbf{D}_f = \{ \mathcal{T} \}^n$. Each tuple $\mathcal{T}=(c_p, c_o)$ is a composition of two concepts, the property $c_p$ and the object $c_o$. 
The environment defines a finite space of $c_p$ and $c_o$. Specifically, we consider the objects as colored tiles in the grid world. Properties are the activated effects when the agent steps onto each tile type. The number of colors for these tiles in a given environment is $n$.  

For this kind of learning to be feasible, the agent is not trained on a single environment, but rather on a group of environments $G_{train} = \{\mathbf{E}_f\}^m$ which varies in $\mathcal{F}$ and subsequently $\mathbf{D}$. 
We assume that the agent has encountered every concept in $G_{train}$, but not every possible $\mathcal{T}$. Zero-shot learning entails understanding new $\mathcal{T}$'s that have not appeared during training. Thus, the agent is evaluated on a group of environments $G_{test}$ that is a superset of $G_{train}$, containing all possible $\mathcal{T}$s.

\vspace{-0.1in}
\section{BabyAI++ platform for zero-shot compositional learning}
\vspace{-3pt}
\label{sec:babyai++}

To facilitate the study of ZSCL, we introduce an augmented platform BabyAI++ based on BabyAI. We first provide an overview of the BabyAI platform before detailing our modifications.

\vspace{-0.1in}
\subsection{BabyAI} %
The BabyAI platform is a configurable platform for procedurally-generated grid-world style environments (based on MiniGrid~\cite{gym_minigrid}) and tasks. Environments in BabyAI consist of grid-world maps with single or multiple rooms. Rooms are randomly populated with objects that can be picked up and dropped by the agent, and doors that can be unlocked and opened by the agent. The number of rooms, objects in the rooms, and connectivity between rooms can be configured and randomized from episode to episode. 
Hence, the BabyAI environment naturally examines the agent's capability of adaption. %
Tasks in BabyAI are procedurally generated from the map, involving going to a specific object, picking up an object, placing an object next to another, or a combination of all three (to be executed in sequence). The task is communicated to the agent through Baby-Language: a compositional language that uses a small subset of English vocabulary. %
Similar to the map, tasks can be randomly generated in each episode based on the selection of objects. While the BabyAI platform incorporates many challenging scenarios such as multi-task learning, partial-observability, and instruction following, it does not involve variable environment dynamics or descriptive text. These features are added by 
BabyAI++.

\vspace{-2mm}
\subsection{BabyAI++}
\textbf{Environment dynamics} \hspace{3pt} In BabyAI++, we add various properties to special floor tiles, each property denotes a different kind of state transition dynamic. For example, stepping on a \textit{"trap"} tile will cause the episode to end with zero reward, and attempting forward movement on a \textit{"flipUpDown"} tile will cause the agent to move backwards. These tiles are distinguished from normal tiles by their color. Similar to objects, the tiles are randomly placed on the map. %

A BabyAI++ level that uses dynamic tiles need to specify parameters such as which types of tiles can appear, how many types of tiles can appear simultaneously, and the frequency by which the tiles appear (in place of a normal tile). The descriptive dynamics of a level can be described by a set of tile colors $\mathcal{C}$ and the set of dynamic properties $\mathcal{P}$. Each instance of the level initializes a new many to one mapping from color to property $\mathcal{M}(c): \mathcal{C} \mapsto \mathcal{P}$, as well as the color mapping of a grid location $\mathcal{G}(x, y): \mathbb{Z}^2 \mapsto \{\mathcal{C}, 0\}$ where $0$ represents no color. 
The following tile properties are currently implemented:
\vspace{-0.1in}
\begin{itemize}[noitemsep]
    \item \textit{trap}: ending the episode with zero reward upon enter;
    \item \textit{slippery}: walking over slippery tile increments the time counter by just a half, thereby increasing reward;
    \item \textit{flipLeftRight}: swapping the left and right actions when the agent is on this tile;
    \item \textit{flipUpDown}: swapping the forward and backward actions when the agent is on this tile;
    \item \textit{sticky}: the agent needs to take 3 actions to leave the tile;
    \item \textit{magic}: if the agent spends more than 1 turn on this tile, it is propelled downward by 1, in addition to the movement of its normal action.
\end{itemize}
\vspace{-0.1in}

\textbf{Descriptive text} \hspace{3pt} Descriptive text about the model dynamics is provided alongside the instructions as observation to the agent. In BabyAI++, we use text to describe which types of tiles are in use and what color is matched to each tile type. Since the pairings between color and tile type are randomized, an agent must learn to ground the descriptive text to navigate efficiently. By default, each color to dynamic pair is described by a sentence in the description, but we also provide more challenging configurations such as partial descriptions. Fig.~\ref{fig:babyai++} provides an example BabyAI++ train/test task.

\textbf{BabyAI++ levels} \hspace{3pt} We build BabyAI++ levels (Table~\ref{tab:env_heldout}) upon BabyAI levels. By enabling different variable dynamics - one BabyAI level can be extended into multiple BabyAI++ levels. 
Note that even simple BabyAI levels can be made challenging with the addition of tiles as the agent must learn to associate language descriptions of tiles to the learned dynamics of each tile type. 

To evaluate language grounding, we partition every level into \emph{training} and \emph{testing} configurations. In the training configuration, the agent is exposed to all tile and colors types in the level, but some combinations of color-type pairs are held-out. For reference, we list complete training configurations in Table~\ref{tab:env_heldout} and report the training successful rate for IL/RL agents.
In the testing configuration, all color-type pairs are enabled. Hence, the agent needs to use language grounding to associate the type of the tile to the color when encountering new color-type pairs at test time. The held-out pairrings are deferred to Appendix.%

\newcommand{\bluecir}{\begin{tikzpicture}[thick,fill opacity=0.5]
\filldraw[fill=blue] (0:1.2mm) circle (1.2mm);
\end{tikzpicture}}
\newcommand{\greencir}{\begin{tikzpicture}[thick,fill opacity=0.5]
\filldraw[fill=green] (0:1.2mm) circle (1.2mm);
\end{tikzpicture}}
\newcommand{\graycir}{\begin{tikzpicture}[thick,fill opacity=0.5]
\filldraw[fill=gray] (0:1.2mm) circle (1.2mm);
\end{tikzpicture}}
\newcommand{\blackcir}{\begin{tikzpicture}[thick,fill opacity=0.5]
\filldraw[fill=black] (0:1.2mm) circle (1.2mm);
\end{tikzpicture}}
\newcommand{\tabcomment}[1]{\multirow{3}{*}{\hspace{-1em}$\left.\begin{array}{l} \\ \\ \\ \end{array}\right\rbrace \text{#1}$}}
\newcommand{\linecomment}[1]{\multirow{1}{*}{\hspace{-1em}$\left.\begin{array}{l} \\ \end{array}  \right. \boldsymbol{\rbrace} \  \text{#1}$}}

\renewcommand{\arraystretch}{1.2}

\begin{table}[htbp!]
    \centering
    \caption{Properties of levels used in experiments (\textit{training} configurations). ``$N$ tiles'' is the number of types of tiles that are simultaneously used in one episode. ``Distractor'' is whether distractor objects are used. The colored circle denotes this property is enabled for specific type of floor tiles in training (e.g., for \texttt{PickupLocal} level, the blue floors can be either \textit{trap} or \textit{flipLeftRight}, and the green floors can be either \textit{slippery} and \textit{flipLeftRight}).}
    \label{tab:env_heldout}
    \resizebox{1.0\linewidth}{!}{
    \begin{tabular}{c|cccc|cccccc|ccl}
        Environment & \rotatebox{90}{Room size} & \rotatebox{90}{$N$ tiles} & \rotatebox{90}{Distractor} & \rotatebox{90}{Var. Target} & \rotatebox{90}{\textit{trap}}&  \rotatebox{90}{\textit{slippery}} &  \rotatebox{90}{\textit{flipLeftRight}} &   \rotatebox{90}{\textit{flipUpDown}} &  \rotatebox{90}{\textit{sticky}} &  \rotatebox{90}{\textit{magic}} & \rotatebox{90}{RL Succ. (\%)} & \rotatebox{90}{IL Succ. (\%)}\\
        \hline
        \texttt{GoToRedBall-v1}  & $1\times1$ & 3 & &  & \bluecir \graycir & \bluecir \greencir & \bluecir \greencir & \bluecir \graycir & \greencir \graycir & \greencir \graycir & 99.7 & 99.2 & \tabcomment{easy goal \& hard ZSCL} \\
        \texttt{GoToRedBall-v2}  & $3\times3$ & 3 & & & \bluecir \graycir & \bluecir \greencir & \bluecir \greencir & \bluecir \graycir & \greencir \graycir & \greencir \graycir & 84.6 & 87.1\\
        \texttt{GoToObj}  & $3\times3$ & 3 &  & \checkmark & \bluecir \graycir & \bluecir \greencir & \bluecir \greencir & \bluecir \graycir & \greencir \graycir & \greencir \graycir & 46.2 & 63.7 & \\
        \texttt{PickupLoc}  & $1\times1$ & 2 & \checkmark & \checkmark & \bluecir &
        \greencir & \bluecir \greencir & -- & -- & --  & 94.9 & 65.8 & \tabcomment{hard goal \& easy ZSCL}\\
        \texttt{PutNextLocal}  & $1\times1$ & 2 & \checkmark & \checkmark & \bluecir & 
        \greencir & \bluecir \greencir & -- & -- & -- & 83.4 & 55.1 & \\
        \texttt{GoTo-v1}  & $3\times3$ & 2 & \checkmark & \checkmark & \bluecir & 
        \greencir & \bluecir \greencir & -- & -- & -- & 57.5 & 63.5 & \\
        \texttt{GoTo-v2}  & $3\times3$ & 3 & \checkmark & \checkmark & \bluecir \graycir & \bluecir \greencir & \bluecir \greencir & \bluecir \graycir & \greencir \graycir & \greencir \graycir & 40.5 & 50.2 & \linecomment{hard goal \& hard ZSCL} \\
    \end{tabular}
    }
\end{table}
\renewcommand{\arraystretch}{1.0}

 \vspace{-0.1in}
\section{Zero-shot compositional policy learning via language grounding}
\label{sec:method}
\vspace{-2pt}
To leverage descriptive languages in policy learning (PL), we first adapt two multi-modal baselines from language-conditional PL. However, existing techniques cannot deal with the zero-shot setting since the agents fail to generalize to unseen environments by only conditioning the observed vision-dynamics pairings. Therefore, we further introduce an attention mechanism to calibrate visual-language representations to assist the agent figure out the correspondence of different attributes.

\begin{figure*}[t]
\centering
\includegraphics[width=0.85\textwidth]{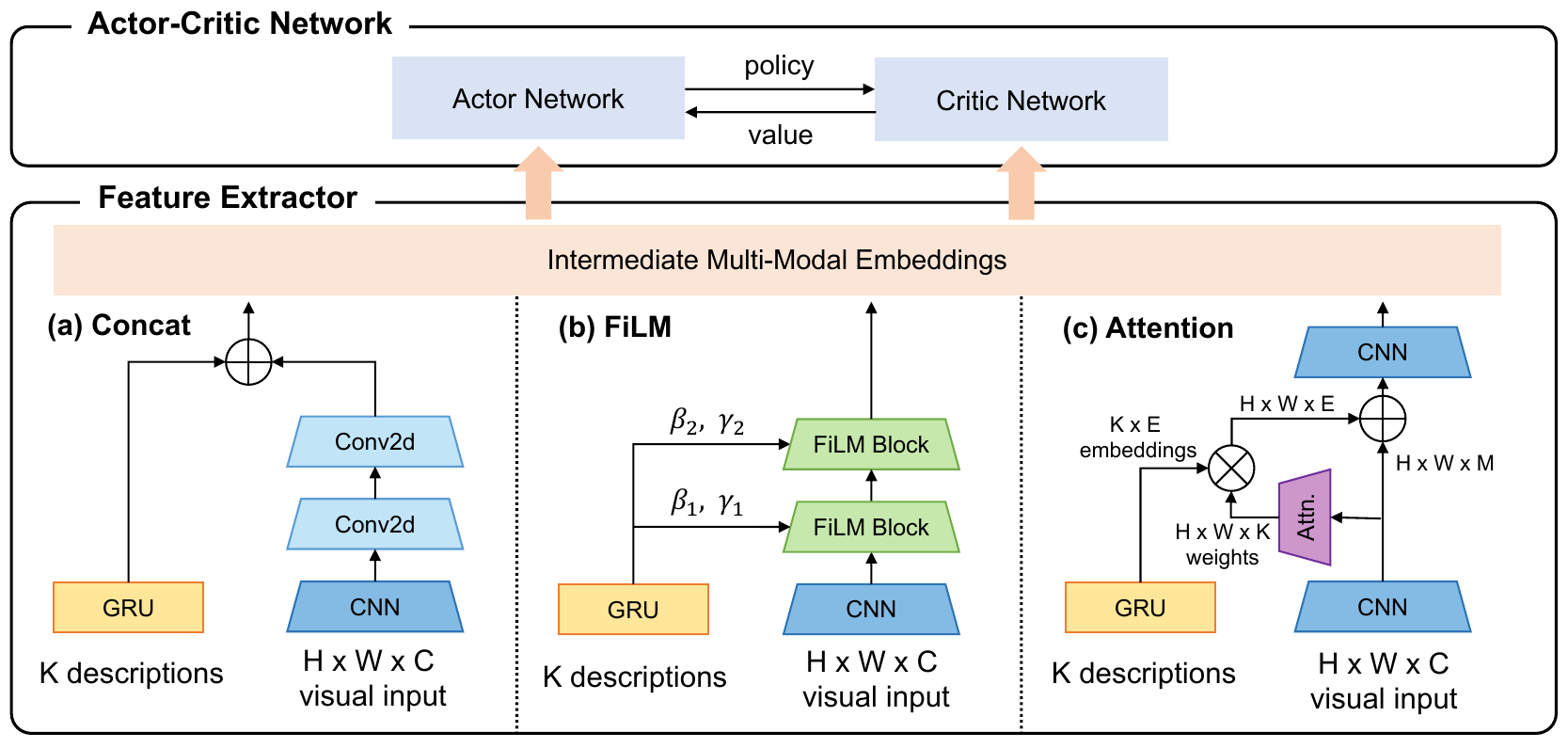}
\vspace{-2mm}
\caption{Comparison of three multi-modal fusion methods: (a) \textit{concat}, (b) \textit{FiLM}~\cite{FiLM} and (c) \textit{attention}. Our method introduces the attention mechanism and assigns language embeddings to visual input explicitly to ground language during training. $H$, $W$, $C$ denotes the height, width, channel of the input images. $E$ represents the dimensionality of descriptive embeddings. $M$ represents the number of the output channels of CNN.}
\vspace{-0.2in}
\label{fig:methods}
\end{figure*}

\vspace{-0.1in}
\subsection{Multi-modal policy learning baselines}
\label{sec:baselines}
\vspace{-2pt}
In our ``PL + language'' setting, the agents need to combine language descriptions and visual inputs. A straight forward pipeline for dealing with multi-modal data is to extract and fuse the features of multi-modal data at different levels. Depending on the task and types of inputs, the researchers might consider sharing the \textit{feature extractor} (backbone) or branch out with independent networks. %
In what follows, we present two baselines inspired by multi-modal literature and language-conditional PL.
\vspace{-0.1in}
\paragraph{Concat-fusion} First, we consider using independent networks to obtain scene embeddings and text embeddings separately. Specifically, we utilize GRU~\cite{GRU} and CNN networks to encode the descriptions and visual scenes.
Then, the scene/text embeddings are concatenated to serve as representations for policy learning algorithms. In this paper, we use the actor-critic network and PPO algorithm~\cite{ppo} for RL and behavior cloning for IL.
\vspace{-0.1in}
\paragraph{FiLM-fusion} 
Inspired by language-conditioned RL works~\cite{babyai}, another baseline we considered is using FiLM layers~\cite{FiLM} to inform the text information. %
Similarly, this technique uses GRU and CNN networks to encode the text and scene embeddings respectively. Then a feature-wise affine transformation is computed based on text embeddings and then imposed to the scenes embeddings:
\begin{align*}
    FiLM (\mathbf{F}_{c} | \gamma_{c} , \beta_{c}) = \gamma_{c} \cdot \mathbf{F}_{c} + \beta_{c}, \text{where} \ \gamma_c = f_c(\mathbf{d}), \beta_{c} = h_c(\mathbf{d}).
\end{align*}
Here, $\mathbf{F}_c$ denotes the $c$-th feature at intermediate layers; $\mathbf{d}$ is the text embeddings; $f, h$ are learnable functions in FiLM layers; the affine transformation parameters $\gamma, \beta$ are the outputs of FiLM layer. Finally, the text-conditioned scene embeddings are fed into the PL algorithms as the observations.
\vspace{-2pt}
\subsection{Calibrating visual-language representations using attention mechanism}
\label{sec:attention_model}
\vspace{-2pt}
Even though multi-modal fusion and conditioning methods can incorporate language priors, leading to improvements during training, they still fail to generalize to unseen vision-dynamics pairings in zero-shot test environments. This is because these approaches do not learn to ground language, thus cannot understand the relationships between vision attributes and environment dynamics through descriptive texts. %

\vspace{-0.1in}
\paragraph{Multi-modal attention-fusion (MMA)} Instead of conditioning the feature maps by texts, we introduce an attention mechanism to assign the text embeddings to every location of the scene embedding feature maps (see (c) in Figure \ref{fig:methods}). This takes advantage of the structured grid-world environment to explicitly ground descriptions of the environment with the observed scene by predicting which description sentence should be assigned to each floor. 

First, for each description sentence $s_i$, we compute the description embedding $\mathbf{d}_i$, and concatenate these text embeddings to form a description dictionary $\mathbf{d}$. Then, the attention module processes the visual image $\mathbf{x}$ and outputs the ``attention-probabilities'' with a size of $H \times W \times (k + 1)$ , where $k$ is the number of descriptive sentences and the last dimension allows zero weight, that is, binding no descriptions to some visual location.
These attention weights are then normalized by the softmax function to obtain an affine transformation of the description embeddings $d_i$. Finally, the transformed description embeddings are spatially concatenated to each floor in the scene embedding to obtain the final compound embeddings. The procedure of attention-fusion model is depicted as follows:
$$\mathbf{F}^{final} = \text{NN}_{fuse}\left(\left[\text{NN}_{feat}\left(\mathbf{x}\right); \text{NN}_{att}(\mathbf{x}) \circ \mathbf{d} \right]\right),$$
where $\text{NN}_{feat}$ and $\text{NN}_{att}$ are two networks for extracting features from images and assigning text attentions; $\text{NN}_{fuse}$ generates compound embeddings $\mathbf{F}^{final}$ by fusing the multi-modal features. %

\vspace{-1mm}
\section{Experiments}
\label{sec:demonstration}
\vspace{-1mm}
\begin{figure}
    \centering
    \includegraphics[width=\textwidth]{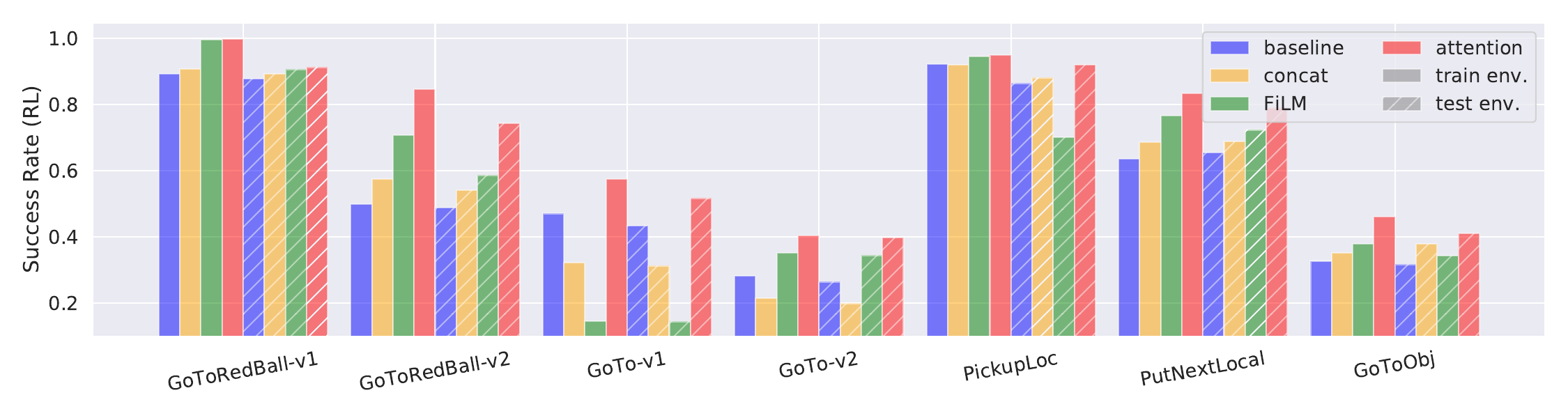}
    \includegraphics[width=\textwidth]{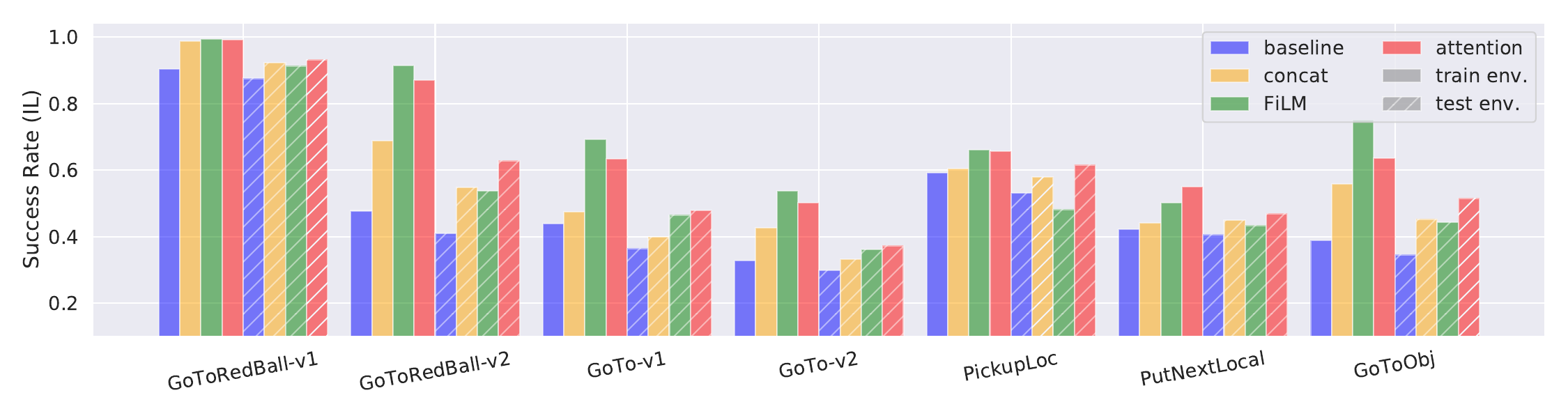}
    \vspace{-0.15in}
    \caption{Comparison of proposed models on seven BabyAI++ environments for RL (top) and IL (bottom). Note that the zero-shot testing results are presented in dashed bars.}
    \label{fig:main_exps}
    
\end{figure}

We thoroughly evaluate the proposed zero-shot compositional policy learning task on varied environments in the BabyAI++ platform. Experiments show that our approach leads to substantial improvements over other multi-modal fusion baselines consistently. Due to the space limit, we defer the ablation studies, implementation details, and more visualizations in the supplementary material.

\vspace{-2mm}
\subsection{Experimental setup}\vspace{-2mm}
\label{ap:train_test_config}

\paragraph{Environments with descriptive text} We conduct extensive evaluations on seven environments on BabyAI++ (see Figure~\ref{fig:main_exps}), which encompass a large variety of tasks in diverse difficulty levels. For instance, \texttt{GoToRedBall-v2} has a larger room size ($3 \times 3$) while less visual attributes and goals than \texttt{GoTo-v2}. For each environment, a subset of visual-dynamics pairings are held out during training, and all the combinations are activated during testing. %

\vspace{-0.1in}
\paragraph{Baselines}
We first take the ``image-only'' model that ignores the descriptive language as the baseline. Then three multi-modal fusion methods introduced in Sec~\ref{sec:method} are evaluated, including \textit{concat-fusion}, \textit{FiLM-fusion}~\cite{FiLM,babyai} and multi-modal attention-fusion. For a fair comparison, we trained all models with the same configurations as~\cite{babyai}. Concretely, we used the PPO algorithm for RL experiments and entropy-regularized behavioral cloning for IL experiments. In all experiments, the Adam optimizer was adopted with the learning rate of $1e^{-4}$. We adopted different number of episodes for different environments according to the task difficulty. Specifically, the RL agents are trained for 10M, 100M, 100M frames for \texttt{GoToRedBall-v1}, \texttt{PickupLoc} and \texttt{PutNextLocal}. For the other environments, the RL agents are trained for 200M frames. As for IL experiments, the agents are trained 20 episodes for $1\times1$ room or 40 episodes for $3\times3$ room with 1M demonstrations.

\begin{table*}[t]
\centering
\caption{Comparison of proposed models on \texttt{PickupLoc} and \texttt{GoToRedBall-v2}. Succ. and $\mathcal{R}_{avg}$ denote the success rate and average reward, the higher the better. $N_{epi}$ denotes the average steps taken in each episode and is only used as reference for the task difficulty. %
The performance is evaluated and averaged for 1,000 episodes on training and testing configurations.} %
\vspace{-0.1in}
\label{tab:main_experiment}
\resizebox{1.0\textwidth}{!}{
\begin{tabular}{@{}cccccccc@{}}
\toprule
\multicolumn{2}{c}{\multirow{2}{*}{Model}} & \multicolumn{3}{c}{Training} & \multicolumn{3}{c}{Testing} \\ \cmidrule(l){3-8} 
\multicolumn{2}{c}{} & Succ. & $\mathcal{R}_{avg}$ & $N_{epi}$ & Succ. & $\mathcal{R}_{avg}$ & $N_{epi}$  \\ \midrule
\rowcolor{gray!40}
\multicolumn{8}{l}{\textit{\textbf{PickupLoc}}: $1\times1$ room, two colors (kind of tiles), three environment dynamics} \\
\multirow{4}{*}{RL} & baseline &  $0.921\pm0.009$ & $0.643\pm0.009$ & $23.755\pm0.547$ & $0.863\pm0.011$ & $0.603\pm0.010$ & $23.655\pm0.558$ \\ 
& concat & $0.920\pm0.009$ & $0.664\pm0.009$ & $22.763\pm0.570$ & $0.880\pm0.010$ & $0.626\pm0.010$ & $22.762\pm0.563$ \\
& FiLM~\cite{FiLM} & $\boldsymbol{0.944\pm0.007}$ & $\boldsymbol{0.679\pm0.008}$ & $22.410\pm0.543$ & $0.701\pm0.014$ & $0.514\pm0.012$ & $29.260\pm0.735$ \\
& MMA (ours) & $\boldsymbol{0.949\pm0.007}$ & $\boldsymbol{0.682\pm0.008}$ & $22.624\pm0.555$ & $\boldsymbol{0.919\pm0.009}$ & $\boldsymbol{0.656\pm0.009}$ & $22.229\pm0.546$ \\ \midrule

\multirow{4}{*}{IL} & baseline &  $0.592\pm0.016$ & $0.406\pm0.012$ & $35.261\pm0.790$ & $0.531\pm0.016$ & $0.371\pm0.012$ & $34.331\pm0.794$ \\
& concat & $0.604\pm0.015$ & $0.406\pm0.012$ & $35.447\pm0.773$ & $0.579\pm0.016$ & $0.399\pm0.012$ & $33.017\pm0.789$ \\
& FiLM~\cite{FiLM} & $\boldsymbol{0.661\pm0.015}$ & $\boldsymbol{0.441\pm0.012}$ & $35.971\pm0.775$ & $0.482\pm0.016$ & $0.334\pm0.012$ & $38.503\pm0.803$ \\
& MMA (ours) & $\boldsymbol{0.658\pm0.015}$ & $\boldsymbol{0.454\pm0.012}$ & $35.877\pm0.788$ & $\boldsymbol{0.616\pm0.015}$ & $\boldsymbol{0.414\pm0.012}$ & $33.694\pm0.771$ \\

\midrule 
\rowcolor{gray!40}
\multicolumn{8}{l}{\textit{\textbf{GoToRedBall-v2}}: $3\times3$ room, three colors (kind of tiles), six environment dynamics} \\
\multirow{4}{*}{RL} & baseline &  $0.499\pm0.016$ & $0.435\pm0.014$ & $227.591\pm7.649$ & $0.488\pm0.016$ & $0.413\pm0.014$ & $239.536\pm7.584$ \\ 
& concat & $0.575\pm0.016$ & $0.501\pm0.014$ & $207.784\pm7.301$ & $0.541\pm0.016$ & $0.465\pm0.014$ & $217.069\pm7.395$ \\
& FiLM~\cite{FiLM} & $0.707\pm0.014$ & $0.587\pm0.013$ & $201.698\pm6.784$ & $0.586\pm0.016$ & $0.490\pm0.014$ & $222.009\pm7.223$ \\
& MMA (ours) & $\boldsymbol{0.846\pm0.011}$ & $\boldsymbol{0.728\pm0.011}$ & $152.310\pm5.920$ & $\boldsymbol{0.743\pm0.014}$ & $\boldsymbol{0.633\pm0.013}$ & $171.782\pm6.194$ \\ \midrule

\multirow{4}{*}{IL} & baseline &  $0.478\pm0.016$ & $0.372\pm0.013$ & $210.403\pm7.175$ & $0.410\pm0.016$ & $0.331\pm0.013$ & $242.642\pm7.736$ \\
& concat & $0.688\pm0.015$ & $0.567\pm0.013$ & $125.479\pm4.832$ & $0.548\pm0.016$ & $0.446\pm0.014$ & $189.918\pm6.908$ \\
& FiLM~\cite{FiLM} & $\boldsymbol{0.914\pm0.009}$ & $\boldsymbol{0.754\pm0.009}$ & $127.810\pm4.170$ & $0.537\pm0.016$ & $0.442\pm0.014$ & $253.688\pm7.749$ \\
& MMA (ours) & $0.871\pm0.011$ & $0.707\pm0.010$ & $147.107\pm4.825$ & $\boldsymbol{0.628\pm0.015}$ & $\boldsymbol{0.516\pm0.013}$ & $219.560\pm6.892$ \\
\bottomrule
\end{tabular}
}
\end{table*}

\vspace{-2mm}
\subsection{Results}
\begin{figure*}[t]
    \centering
    \includegraphics[width=0.32\textwidth]{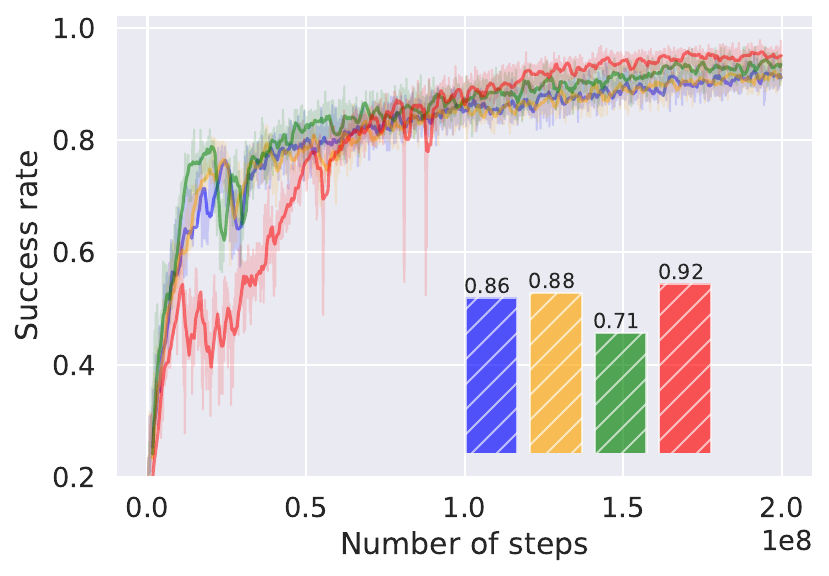}
    \includegraphics[width=0.32\textwidth]{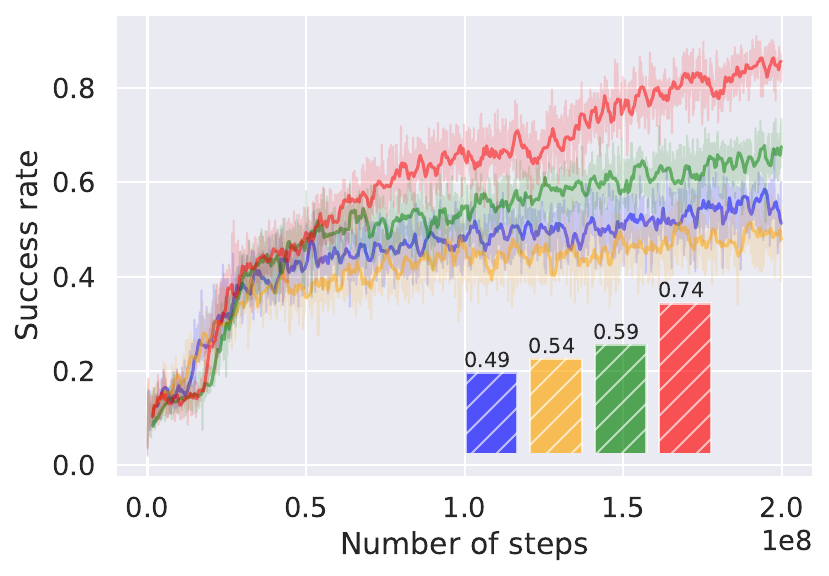}
    \includegraphics[width=0.33\textwidth]{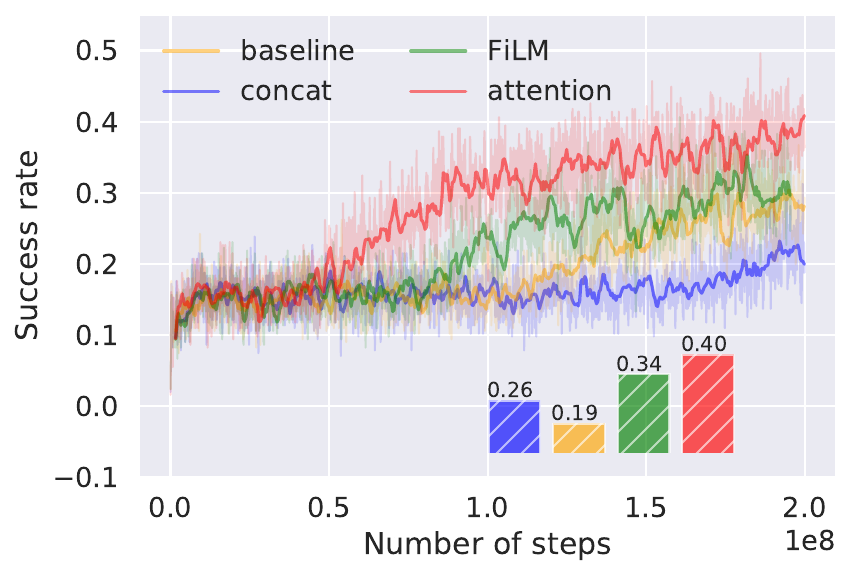}
    \caption{Learning curves of RL algorithms on three training environments: \texttt{PickupLoc} (left), \texttt{GoToRedBall-v2} (middle) and \texttt{GoTo-v2} (right). ``attention'' denotes proposed MMA model in Sec~\ref{sec:attention_model}. \underline{The zero-shot testing results are given in the dashed bars.}} %
    \label{fig:learning_curve}
    \vspace{-0.1in}
\end{figure*}

\begin{figure*}[t]
    \centering
    \includegraphics[width=0.25\textwidth]{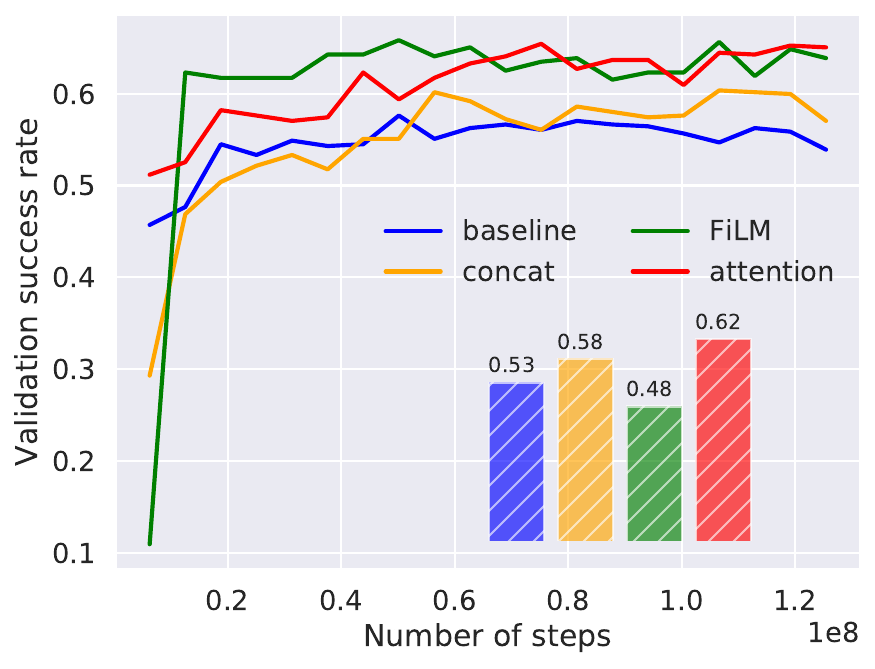}
    \hspace{-2mm}
    \includegraphics[width=0.25\textwidth]{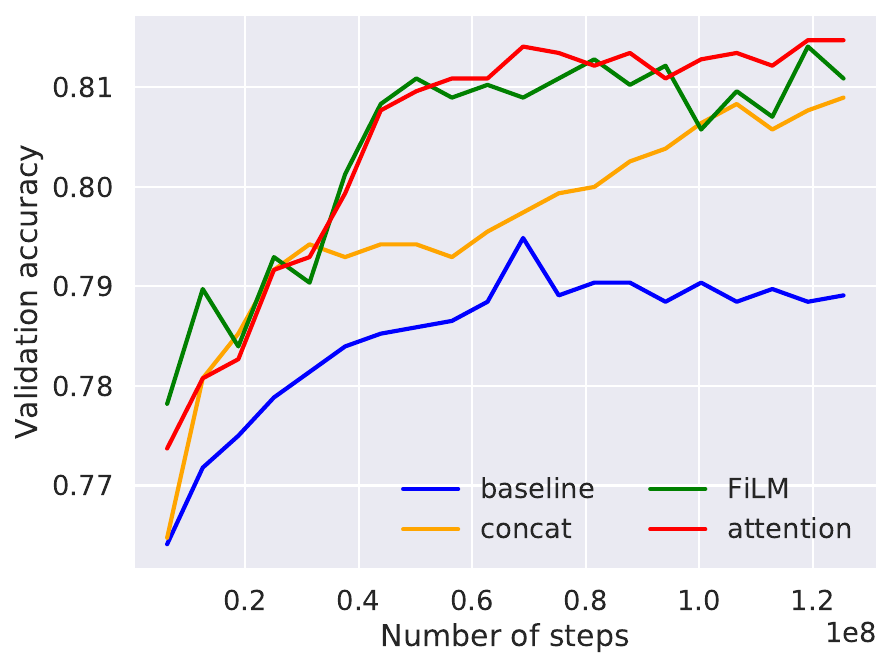}
    \hspace{-2mm}
    \includegraphics[width=0.25\textwidth]{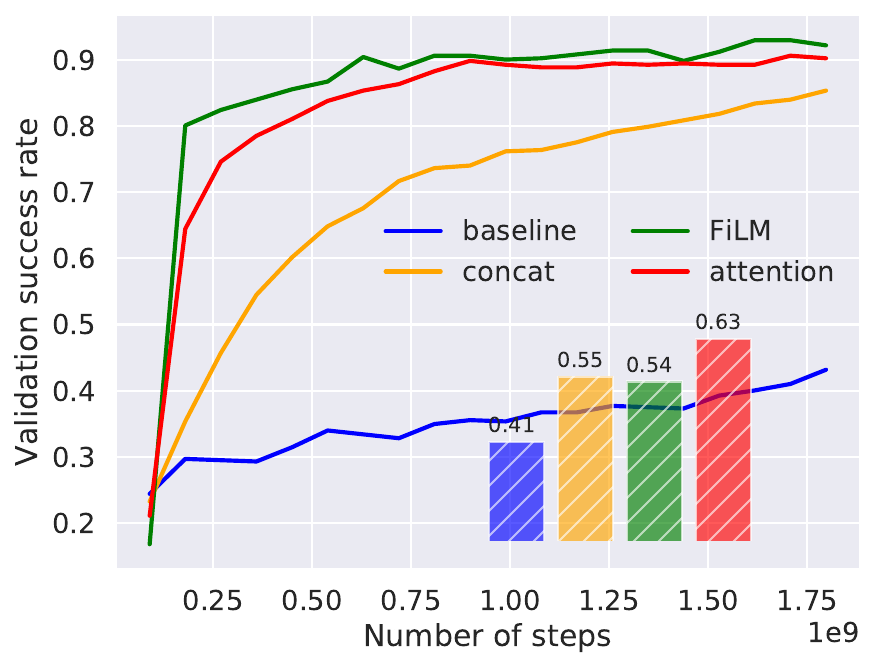}
    \hspace{-2mm}
    \includegraphics[width=0.25\textwidth]{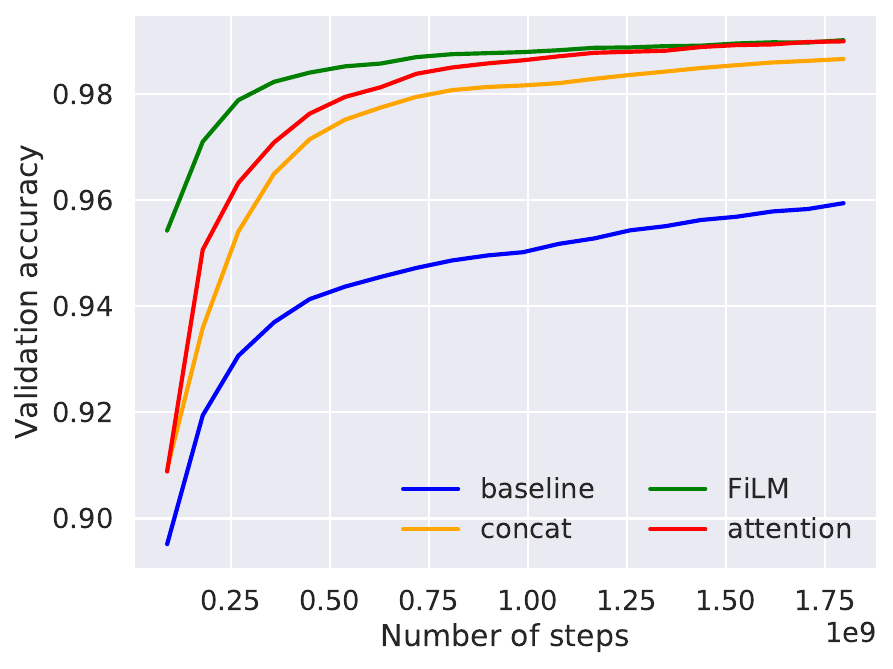} \\
    \hspace{0.5cm} (a) \hspace{3cm} (b) \hspace{3cm} (c) \hspace{3cm} (d)
    \caption{Learning curves of IL algorithms on \texttt{PickupLoc} (a, b) and \texttt{GoToRedBall-v2} (c, d) training environments. \underline{The zero-shot testing results are given in the dashed bars.} The validation success rate and behavior cloning accuracy are reported in top and bottom rows respectively, where the performance is evaluated on 512 validation expert trajectories on \underline{training configurations}.}
    \label{fig:learning_curve_IL}
    \vspace{-0.15in}
\end{figure*}

\vspace{-2mm}
\paragraph{Benchmarking BabyAI++ levels:}
Figure~\ref{fig:main_exps} shows the performance of proposed models for RL tasks on BabyAI++. In general, the \textit{attention-fusion} model achieves the best overall performance on both training and testing environments, and the benefits become more remarkable during testing. On the contrary, while the FiLM model leads to similar performance as our approach during training, it suffers from a significant performance drop (e.g., on \texttt{PickupLoc}) when facing unseen visual-dynamics pairings due to train-test distribution shift. 
Table~\ref{tab:main_experiment} provides quantitative comparisons of other metrics on two selected environments. 
Specifically, our model results in $29.5\%$ and $21.8\%$ improvement (averged on RL and IL) over the FiLM model on \texttt{PickupLoc} and \texttt{GoToRedBall-v2}, while on par with or outperforming FiLM during training.
\vspace{-2mm}
\paragraph{Training curves vs. zero-shot testing results:}
We present the learning curves of RL algorithms in Figure~\ref{fig:learning_curve}. Our approach converges better/faster in more complicated environments compared to other fusion models. It again verifies the effectiveness of the attention mechanism to enforce the agent to ground the language.
Similarly, Figure~\ref{fig:learning_curve_IL} shows the learning curves of IL algorithms on two BabyAI++ environments. Compared to ``image-only'' baseline (blue line), multi-modal fusion models yield a significant improvement in behavior cloning accuracy thus leading to a better success rate. However, there is a huge performance gap in test configurations where the visual-dynamics links are novel. The overfitting phenomenon is remarkable for language-conditional technique FiLM and mitigated by proposed MMA approach. 

\vspace{-2mm}
\paragraph{Learning with instructive language:}
Apart from descriptive text, instructions also play an important role because the targets for more advanced levels (e.g., the object in \texttt{GotoObj}) vary per episode. From Table~\ref{tab:main_experiment} and~\cite{babyai}, we know FiLM is effective in learning conditioned instructions but cannot reason the unseen test configurations through grounding the descriptive text. Therefore, we propose a hybrid model that deploys the attention-fusion model to ground the descriptive language and then utilize FiLM to condition the post-processing module with the task instructions. Table~\ref{tab:ablation_instructions} shows an example that the proposed hybrid model surpasses other models by a large margin.

\begin{table*}[htbp!]
\caption{Comparison of proposed models with different types of texts on \texttt{PutNextLocal}. Succ. and $\mathcal{R}_{avg}$ denote the success rate and average reward, the higher the better. $N_{epi}$ denotes the average steps taken in each episode and is only used as reference for the task difficulty.} Since the goals are altered at each episode, instructions also play a significant role as shown in~\cite{babyai}.
\label{tab:ablation_instructions}
\centering
\resizebox{1.0\textwidth}{!}{
\begin{tabular}{@{}ccccccccc@{}}
\toprule
\multicolumn{3}{c}{\multirow{2}{*}{Model}} & \multicolumn{3}{c}{Training} & \multicolumn{3}{c}{Testing} \\ \cmidrule(l){4-9} 
\multicolumn{3}{c}{} & Succ. & $\mathcal{R}_{avg}$ & $N_{epi}$ & Succ. & $\mathcal{R}_{avg}$ & $N_{epi}$  \\ \midrule

\rowcolor{gray!40}
\multicolumn{9}{l}{\textit{\textbf{PutNextLocal}}: $1\times1$ room, two colors (kind of tiles), three environment dynamics} \\
\multirow{8}{*}{RL} & baseline & no text & $0.636\pm0.015$ & $0.427\pm0.012$ & $64.284\pm1.510$ & $0.654\pm0.015$ & $0.451\pm0.012$ & $58.247\pm1.479$ \\ \cmidrule(l){2-9} 
& \multirow{2}{*}{concat} & descriptive & $0.687\pm0.015$ & $0.450\pm0.011$ & $62.379\pm1.396$ & $0.688\pm0.015$ & $0.442\pm0.011$ & $59.896\pm1.358$ \\
& & all texts & $0.756\pm0.013$ & $0.607\pm0.011$ & $37.197\pm1.132$ & $0.803\pm0.013$ & $0.640\pm0.011$ & $36.771\pm1.139$ \\ \cmidrule(l){2-9} 
& \multirow{2}{*}{FiLM~\cite{FiLM}} & descriptive & $0.766\pm0.013$ & $0.510\pm0.011$ & $58.950\pm1.345$ & $0.723\pm0.014$ & $0.483\pm0.011$ & $59.642\pm1.365$ \\
& & all texts & $0.905\pm0.009$ & $0.784\pm0.008$ & $20.851\pm0.621$ & $0.911\pm0.009$ & $0.789\pm0.008$ & $21.278\pm0.636$ \\ \cmidrule(l){2-9} 
& \multirow{2}{*}{MMA (ours)} & descriptive &  $0.834\pm0.012$ & $0.566\pm0.010$ & $58.754\pm1.278$ & $0.790\pm0.013$ & $0.537\pm0.011$ & $52.591\pm1.261$ \\ 
& & all texts & $\boldsymbol{0.984\pm0.004}$ & $\boldsymbol{0.858\pm0.004}$ & $20.034\pm0.540$ & $\boldsymbol{0.989\pm0.003}$ & $\boldsymbol{0.866\pm0.004}$ & $19.849\pm0.560$ \\
\bottomrule
\end{tabular}
}
\end{table*}

\begin{figure}
    \centering
    \includegraphics[width=\textwidth]{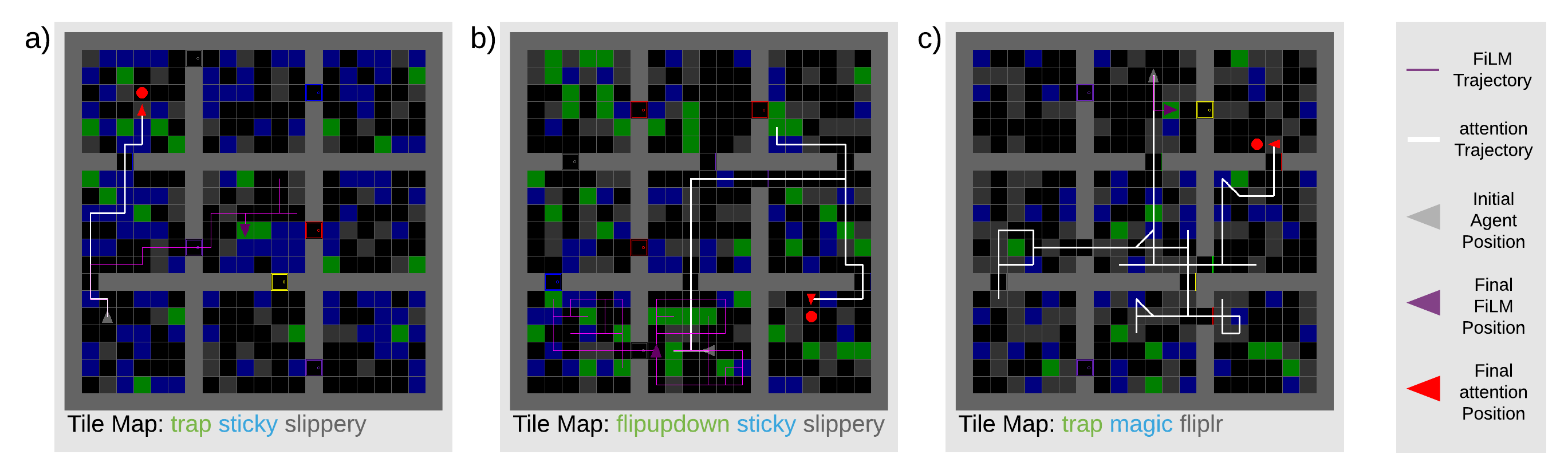}
    \caption{Example trajectories of the proposed attention (white line) and FiLM (purple line) models on \texttt{GoToRedBall-v2}. %
    (a) Our approach avoids the previously unseen green-trap combination while FiLM does not. (b) When the green tile stands for \textit{flipupdown}, our model %
    realizes the transformation and does not avoid it. (c) Our method avoids stepping on the left most green, and leverages the \textit{magic} blue tile properties to hop over the middle and rightmost green traps before arriving at the target.}
    \label{fig:trajectories}
    \vspace{-0.15in}
\end{figure}

\begin{figure}[!ht]
    \centering
    \includegraphics[width=\linewidth]{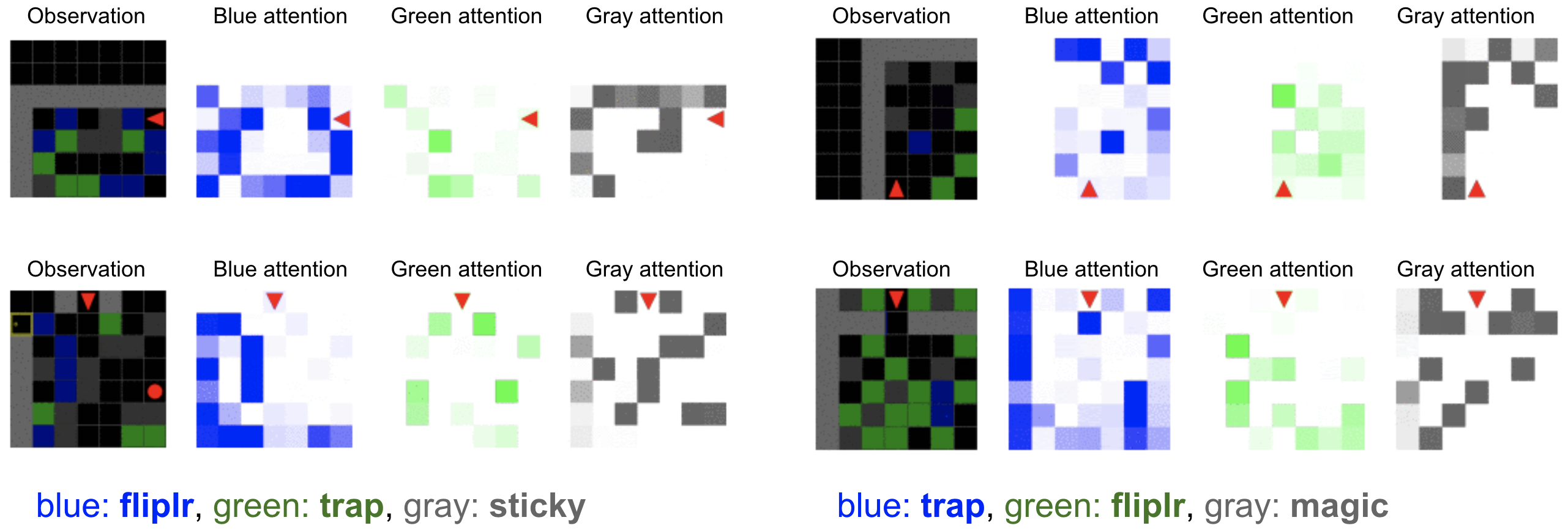}
    \vspace{-0.2in}
    \caption{Visualization of attention module in learned networks trained with RL on \texttt{GotoRedBall-v2}. The first panel shows the observation from the agent's view, and the rest three panels from the right shows the attention weighting to grid positions for each sentence of description. The position of the agent in each panel is highlighted in red. The visual-dyanmics pairings are provided in the bottom.}
    \label{fig:attention}
    \vspace{-0.15in}
\end{figure}

\vspace{-2mm}
\subsection{Visualizing language grounding}\vspace{-2mm}
To verify whether performance improvements of the proposed method was due to language grounding, we first track the trajectories of different approaches on unseen test environments. For instance, as shown in Figure \ref{fig:trajectories}, our model is capable of adjusting its behavior according to the properties of colored tiles on \texttt{GoToRedBall-v2}.
Then we visualize the attention of each sentence in the description to spatial locations in the observation in Figure \ref{fig:attention}. the attention module assigns a larger weight when the tiles are related to corresponding colors. Moreover, the gray attention also automatically captures the walls and doors in addition to the corresponding gray tiles. More visualization examples are deferred to Sec~\ref{sec:attn_viz} in the supplementary material.

\vspace{-2mm}
\section{Conclusion}
\vspace{-2mm}
\label{sec:conclusion}
Generalization to unseen environments is still one of the most challenging problems that hinder the wide applications of policy learning techniques. In this paper, we introduce a challenging zero-shot composition learning task in RL/IL settings by grounding natural language. To spur future research, we develop a generalized BabyAI++ platform to generate a series of grid-world tasks, where an environment's visual appearance and dynamics are disentangled. Moreover, accompanying descriptive texts are generated to capture correspondences between vision attributes and physical dynamics. %
As a first step towards ZSCL in policy learning, the controlled simulation environment BabyAI++ is valuable and challenging, while the current implementation on the grid world has limited capacity for simulating real world environments.
Nonetheless, the evaluation results in this work articulate that language grounding can aid the policy learning agents to learn compositional concepts in a zero-shot manner. We hope our BabyAI++ platform will stimulate future research on policy learning approaches for zero-shot compositional RL and IL problems.

\appendix
\setcounter{section}{0}
\setcounter{figure}{0}
\setcounter{table}{0}

\makeatletter 
\renewcommand{\thefigure}{A\arabic{figure}}
\setcounter{table}{0}
\renewcommand{\thetable}{A\arabic{table}}

\newpage
\section{Experimental Setup}

\subsection{Zero-Shot Configurations}
\label{ap:zero_shot_config}
Since the dynamics of environments in BabyAI++ are disentangled from the visual appearance, it is possible to create novel test configurations for zero-shot compositional learning. Specifically, only partial color-dynamics combinations are activated during training while the other pairs are held-out for testing. Therefore, the agents are required to learn the language grounding to generalize to unseen configurations.
The held-out pairs are listed below:
\begin{itemize}
    \item \texttt{GoToRedBall-v1}, \texttt{GotoRedBall-v2}, \texttt{GotoObj}, and \texttt{Goto-v2}: (\textit{blue}, \textit{sticky}), (\textit{blue}, \textit{magic}), (\textit{green}, \textit{trap}), (\textit{green}, \textit{flipUpDown}), (\textit{grey}, \textit{flipLeftRight}), (\textit{grey}, \textit{slippery}).
    \item \texttt{PutNextLocal}, \texttt{Goto-v1}, and \texttt{PickupLoc}: (\textit{blue}, \textit{slippery}), (\textit{green}, \textit{trap}).
\end{itemize}

\subsection{Implementation Details}
\paragraph{Environments:} We built the environments on top of BabyAI levels. Specifically, we sampled one property (with replacement) for each color that appears in the level when initializing the environments at each episode. The description text is then generated using a fixed sentence structure of ``$\{$\texttt{color}$\}$ tiles are $\{$\texttt{property}$\}$". The placement of tiles in the map is random. Each eligible tile in the map can take on one of the selected properties in the current episode. Different tiles have different chances of being selected. \texttt{trap} has $5\%$ chance of spawn, \texttt{slippery} and \texttt{magic} have $7.5\%$ chance of spawn, and the remaining three dynamics has $10\%$ chance of spawn. Note that this is configured to reduce the likelihood of having path to objects be completed blocked by \texttt{trap} tiles during training.

\paragraph{RL/IL Agents:} 
We adopt the same model architecture following~\cite{babyai} for the \textbf{image-only} baseline and two multi-modal fusion models: \textbf{concat} and \textbf{FiLM}~\cite{FiLM} except that the instructive language is replaced by descriptive text generated by BabyAI++.
For our proposed \textbf{attention}-fusion approach, we keep the feature extractors for visual input (CNNs) and descriptive text (GRU) the same as the other baselines. In what follows, we give the details of attention-fustion architecture.

Specifically, a two-layer convolutional network is used to process the $7\times 7$ visual input, with the kernel size of 3 and the hidden dimension of 128.
For the description text, we split it by sentence and then processed by an embedding layer followed by a GRU layer, both with dimensions of 64 to create representation for each sentence. 
To calibrate the language embeddings in the spatial feature map, we utilize an attention network (also convolutional) with softmax activation to predict the probability of the relation between specific tiles and descriptive sentences, taking the output of the visual CNN.
Then the final language representation of each spatial position is computed by the sum of GRU outputs as weighted by the attention head outputs, which is then concatenated with the scene feature maps. Finally, the multi-modal feature maps are passed through two convolutional layers (128 hidden dimension), and sent to the actor-critic networks for policy learning (two layer fully connected networks, with tanh hidden layer of width 64).

\paragraph{Training and Testing:}
We used the PPO algorithm for RL experiments and entropy-regularized behavioral cloning for IL experiments. In all experiments, the Adam optimizer was adopted with the learning rate of $1e^{-4}$. We adopted different number of episodes for different environments according to the task difficulty. Specifically, the RL agents are trained for 10M, 100M, 100M frames for \texttt{GoToRedBall-v1}, \texttt{PickupLoc} and \texttt{PutNextLocal}. For the other environments, the RL agents are trained for 200M frames. As for IL experiments, the agents are trained 20 episodes for $1\times1$ room or 40 episodes for $3\times3$ room with 1M demonstrations.

\section{Supplementary Results}

\subsection{Quantitative Policy Learning Results}

Figure~\ref{fig:learning_curve_IL} shows the learning curves of IL algorithms on two BabyAI++ environments. Compared to ``image-only'' baseline (blue line), multi-modal fusion models yield a significant improvement in behavior cloning accuracy thus leading to a better success rate. However, there is a huge performance gap in test configurations where the visual-dynamics links are novel. The overfitting phenomenon is remarkable for language-conditional technique FiLM and mitigated by our attention-fusion proposal.

In Table~\ref{tab:main_experiment_supp}, we present the complete results of Table~\ref{tab:main_experiment} on the other five environments. Similarly, our approach outperforms the other baselines consistently during testing while achieving competitive or even the best performance in the training environments. On the contrary, FiLM-fusion approach overfits to the training environments thus resulting in a significant performance drop when facing the novel test configurations.

\begin{table*}[htbp!]
\centering
\resizebox{1.0\textwidth}{!}{
\begin{tabular}{@{}cccccccc@{}}
\toprule
\multicolumn{2}{c}{\multirow{2}{*}{Model}} & \multicolumn{3}{c}{Training} & \multicolumn{3}{c}{Testing} \\ \cmidrule(l){3-8} 
\multicolumn{2}{c}{} & Succ. & $\mathcal{R}_{avg}$ & $N_{epi}$ & Succ. & $\mathcal{R}_{avg}$ & $N_{epi}$  \\ \midrule
\rowcolor{gray!40}
\multicolumn{8}{l}{\textit{\textbf{GoToRedBall-v1}}: $1\times1$ room, three colors (kind of tiles), six environment dynamics} \\
\multirow{4}{*}{RL} & baseline & $0.893\pm0.010$ & $0.783\pm0.009$ & $12.671\pm0.510$ & $0.877\pm0.010$ & $0.770\pm0.010$ & $13.310\pm0.554$ \\
& concat & $0.907\pm0.009$ & $0.798\pm0.009$ & $11.480\pm0.458$ & $0.892\pm0.010$ & $0.778\pm0.009$ & $12.806\pm0.518$ \\
& FiLM~\cite{FiLM} & $\boldsymbol{0.996\pm0.002}$ & $\boldsymbol{0.914\pm0.002}$ & $6.031\pm0.132$ & $\boldsymbol{0.905\pm0.009}$ & $\boldsymbol{0.814\pm0.009}$ & $11.523\pm0.555$ \\
& MMA (ours) & $\boldsymbol{0.997\pm0.002}$ & $\boldsymbol{0.914\pm0.002}$ & $6.003\pm0.124$ & $\boldsymbol{0.912\pm0.009}$ & $\boldsymbol{0.807\pm0.009}$ & $11.818\pm0.515$ \\ \midrule

\multirow{4}{*}{IL} & baseline &  $0.905\pm0.009$ & $0.805\pm0.009$ & $10.146\pm0.441$ & $0.876\pm0.010$ & $0.778\pm0.010$ & $11.822\pm0.525$ \\
& concat & $0.988\pm0.003$ & $0.907\pm0.004$ & $6.020\pm0.139$ & $0.923\pm0.008$ & $\boldsymbol{0.848\pm0.008}$ & $8.867\pm0.401$ \\
& FiLM~\cite{FiLM} & $\boldsymbol{0.994\pm0.002}$ & $\boldsymbol{0.913\pm0.003}$ & $6.000\pm0.142$ & $0.913\pm0.009$ & $0.825\pm0.009$ & $10.640\pm0.497$ \\
& MMA (ours) & $\boldsymbol{0.992\pm0.003}$ & $\boldsymbol{0.910\pm0.003}$ & $6.224\pm0.176$ & $\boldsymbol{0.932\pm0.008}$ & $0.814\pm0.008$ & $11.375\pm0.473$ \\
\midrule 

\rowcolor{gray!40}
\multicolumn{8}{l}{\textit{\textbf{GoToObj}}: $3\times3$ room, three colors (kind of tiles), six environment dynamics} \\
\multirow{4}{*}{RL} & baseline & $0.328\pm0.015$ & $0.284\pm0.013$ & $471.026\pm15.481$ & $0.317\pm0.015$ & $0.274\pm0.013$ & $502.067\pm15.806$ \\
& concat & $0.353\pm0.015$ & $0.300\pm0.013$ & $463.263\pm14.901$ & $0.379\pm0.015$ & $0.329\pm0.014$ & $452.835\pm15.093$ \\
& FiLM~\cite{FiLM} & $0.380\pm0.015$ & $0.314\pm0.013$ & $533.636\pm14.997$ & $0.343\pm0.015$ & $0.282\pm0.013$ & $543.757\pm15.399$ \\
& MMA (ours) & $\boldsymbol{0.462\pm0.016}$ & $\boldsymbol{0.382\pm0.014}$ & $554.729\pm15.080$ & $\boldsymbol{0.411\pm0.016}$ & $\boldsymbol{0.347\pm0.014}$ & $552.472\pm15.626$ \\ \midrule

\multirow{4}{*}{IL} & baseline &  $0.390\pm0.015$ & $0.311\pm0.013$ & $386.501\pm14.332$ & $0.346\pm0.015$ & $0.268\pm0.012$ & $450.040\pm15.191$ \\
& concat & $0.559\pm0.016$ & $0.443\pm0.013$ & $276.233\pm11.096$ & $0.452\pm0.016$ & $0.360\pm0.013$ & $374.064\pm13.653$ \\
& FiLM~\cite{FiLM} & $\boldsymbol{0.752\pm0.014}$ & $\boldsymbol{0.595\pm0.012}$ & $320.308\pm10.613$ & $0.443\pm0.016$ & $0.347\pm0.013$ & $483.975\pm14.707$ \\
& MMA (ours) & $0.637\pm0.015$ & $0.495\pm0.013$ & $419.394\pm12.681$ & $\boldsymbol{0.515\pm0.016}$ & $\boldsymbol{0.396\pm0.013}$ & $472.688\pm13.555$ \\
\midrule

\rowcolor{gray!40}
\multicolumn{8}{l}{\textit{\textbf{PutNextLocal}}: $1\times1$ room, two colors (kind of tiles), three environment dynamics} \\
\multirow{4}{*}{RL} & baseline & $0.636\pm0.015$ & $0.427\pm0.012$ & $64.284\pm1.510$ & $0.654\pm0.015$ & $0.451\pm0.012$ & $58.247\pm1.479$ \\
& concat & $0.687\pm0.015$ & $0.450\pm0.011$ & $62.379\pm1.396$ & $0.688\pm0.015$ & $0.442\pm0.011$ & $59.896\pm1.358$ \\
& FiLM~\cite{FiLM} & $0.766\pm0.013$ & $0.510\pm0.011$ & $58.950\pm1.345$ & $0.723\pm0.014$ & $0.483\pm0.011$ & $59.642\pm1.365$ \\
& MMA (ours) & $\boldsymbol{0.834\pm0.012}$ & $\boldsymbol{0.566\pm0.010}$ & $58.754\pm1.278$ & $\boldsymbol{0.790\pm0.013}$ & $\boldsymbol{0.537\pm0.011}$ & $52.591\pm1.261$ \\ \midrule

\multirow{4}{*}{IL} & baseline &  $0.423\pm0.016$ & $0.290\pm0.012$ & $71.950\pm1.657$ & $0.406\pm0.016$ & $0.258\pm0.011$ & $71.943\pm1.661$ \\
& concat & $0.442\pm0.016$ & $0.293\pm0.012$ & $80.786\pm1.630$ & $0.449\pm0.016$ & $\boldsymbol{0.303\pm0.012}$ & $70.758\pm1.639$ \\
& FiLM~\cite{FiLM} & $0.502\pm0.016$ & $0.330\pm0.012$ & $86.245\pm1.546$ & $0.434\pm0.016$ & $0.277\pm0.011$ & $82.083\pm1.588$ \\
& MMA (ours) & $\boldsymbol{0.551\pm0.016}$ & $\boldsymbol{0.351\pm0.012}$ & $86.864\pm1.536$ & $\boldsymbol{0.469\pm0.016}$ & $\boldsymbol{0.301\pm0.011}$ & $78.643\pm1.575$ \\
\midrule

\rowcolor{gray!40}
\multicolumn{8}{l}{\textit{\textbf{GoTo-v1}}: $3\times3$ room, two colors (kind of tiles), three environment dynamics} \\
\multirow{4}{*}{RL} & baseline &  $0.471\pm0.016$ & $0.399\pm0.014$ & $258.336\pm11.993$ & $0.434\pm0.016$ & $0.366\pm0.014$ & $222.980\pm11.310$ \\
& concat & $0.323\pm0.015$ & $0.273\pm0.013$ & $448.914\pm15.375$ & $0.313\pm0.015$ & $0.261\pm0.013$ & $421.242\pm14.944$ \\
& FiLM~\cite{FiLM} & $0.146\pm0.011$ & $0.133\pm0.010$ & $592.853\pm16.769$ & $0.145\pm0.011$ & $0.130\pm0.010$ & $498.147\pm16.646$ \\
& MMA (ours) & $\boldsymbol{0.575\pm0.016}$ & $\boldsymbol{0.487\pm0.014}$ & $249.288\pm10.760$ & $\boldsymbol{0.517\pm0.016}$ & $\boldsymbol{0.449\pm0.014}$ & $198.593\pm9.661$ \\\midrule

\multirow{4}{*}{IL} & baseline & $0.440\pm0.016$ & $0.332\pm0.013$ & $350.262\pm12.570$ & $0.364\pm0.015$ & $0.268\pm0.012$ & $300.200\pm12.412$ \\
& concat & $0.475\pm0.016$ & $0.350\pm0.013$ & $381.580\pm12.380$ & $0.399\pm0.015$ & $0.300\pm0.013$ & $312.183\pm12.150$ \\
& FiLM~\cite{FiLM} & $\boldsymbol{0.692\pm0.015}$ & $\boldsymbol{0.512\pm0.012}$ & $527.769\pm13.313$ & $0.465\pm0.016$ & $0.327\pm0.012$ & $445.229\pm13.752$ \\
& MMA (ours) & $0.635\pm0.015$ & $0.448\pm0.012$ & $545.996\pm13.083$ & $\boldsymbol{0.479\pm0.016}$ & $\boldsymbol{0.331\pm0.012}$ & $394.302\pm13.372$ \\ \midrule

\rowcolor{gray!40}
\multicolumn{8}{l}{\textit{\textbf{GoTo-v2}}: $3\times3$ room, three colors (kind of tiles), six environment dynamics} \\
\multirow{4}{*}{RL} & baseline &  $0.283\pm0.014$ & $0.233\pm0.012$ & $544.693\pm15.755$ & $0.264\pm0.014$ & $0.226\pm0.012$ & $549.965\pm15.933$ \\
& concat & $0.216\pm0.013$ & $0.187\pm0.012$ & $596.525\pm15.912$ & $0.199\pm0.013$ & $0.170\pm0.011$ & $629.078\pm15.835$ \\
& FiLM~\cite{FiLM} & $0.353\pm0.015$ & $0.293\pm0.013$ & $470.907\pm15.083$ & $0.344\pm0.015$ & $0.283\pm0.013$ & $478.038\pm15.211$ \\
& MMA (ours) & $\boldsymbol{0.405\pm0.016}$ & $\boldsymbol{0.345\pm0.014}$ & $493.855\pm15.413$ & $\boldsymbol{0.398\pm0.015}$ & $\boldsymbol{0.337\pm0.014}$ & $472.162\pm15.099$ \\ \midrule

\multirow{4}{*}{IL} & baseline & $0.328\pm0.015$ & $0.238\pm0.012$ & $435.791\pm14.690$ & $0.299\pm0.014$ & $0.216\pm0.011$ & $495.675\pm15.510$ \\
& concat & $0.426\pm0.016$ & $0.303\pm0.012$ & $407.187\pm13.428$ & $0.332\pm0.015$ & $0.232\pm0.011$ & $488.717\pm15.133$ \\
& FiLM~\cite{FiLM} & $\boldsymbol{0.537\pm0.016}$ & $\boldsymbol{0.388\pm0.013}$ & $527.599\pm13.446$ & $0.361\pm0.015$ & $0.258\pm0.012$ & $599.806\pm14.949$ \\
& MMA (ours) & $0.502\pm0.016$ & $0.365\pm0.013$ & $517.113\pm13.426$ & $\boldsymbol{0.373\pm0.015}$ & $\boldsymbol{0.271\pm0.012}$ & $580.234\pm14.339$ \\

\bottomrule
\end{tabular}
}
\caption{Comparison of proposed models on the other five environments. Succ. and $\mathcal{R}_{avg}$ denote the success rate and average reward, the higher the better. $N_{epi}$ denotes the average steps taken in each episode and is only used as reference for the task difficulty. For all metrics, we present the sample mean together with standard error. 
The performance is evaluated and averaged for 1,000 episodes on training and testing configurations.}
\label{tab:main_experiment_supp}
\end{table*}

\subsection{Ablation Study}

\paragraph{Learning with random language:}

To further validate that performance improvements are due to the use of descriptive text rather than larger model capacity, we conducted an ablation study on the texts for the attention model. Specifically, to show the utility of descriptive text as a vessel for knowledge transfer, we generate and replace original inputs with the following nonsensical texts:
\begin{itemize}
    \item \textit{random text}: generate random texts from a pre-defined dictionary that contains the same number of irrelevant words as the descriptive texts in this environment. 
    \item \textit{shuffled text}: shuffle the descriptive texts randomly at each episode. In this case, the context is broken thus the mapping for the color of tiles and their properties are difficult is difficult to learn.%
\end{itemize}

We found that only the models with meaningful descriptive text are able to produce a much better performance in both training and testing configurations. Moreover, we observe that even the random texts could bring marginal benefits by introducing randomness and implicit exploration, which is consistent with previous literature~\cite{branavan2012learning}.

\subsection{Visualizing the Language Grounding - Attention}\label{sec:attn_viz}
\begin{figure}[!ht]
    \centering
    \includegraphics[width=\linewidth]{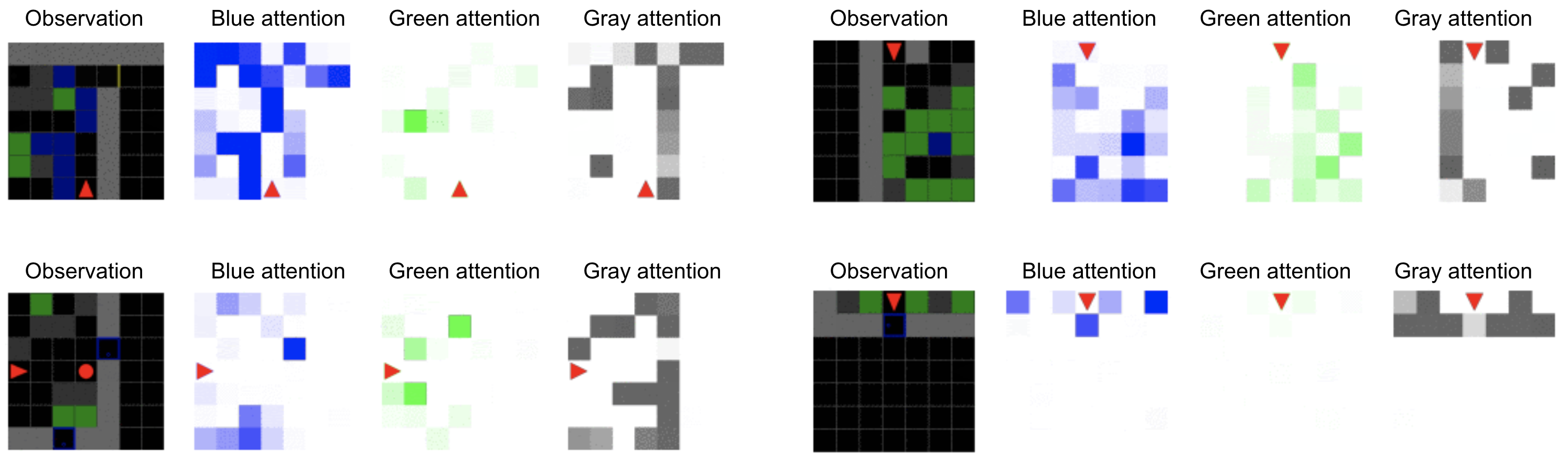} \\ \vspace{3mm}
    \includegraphics[width=\linewidth]{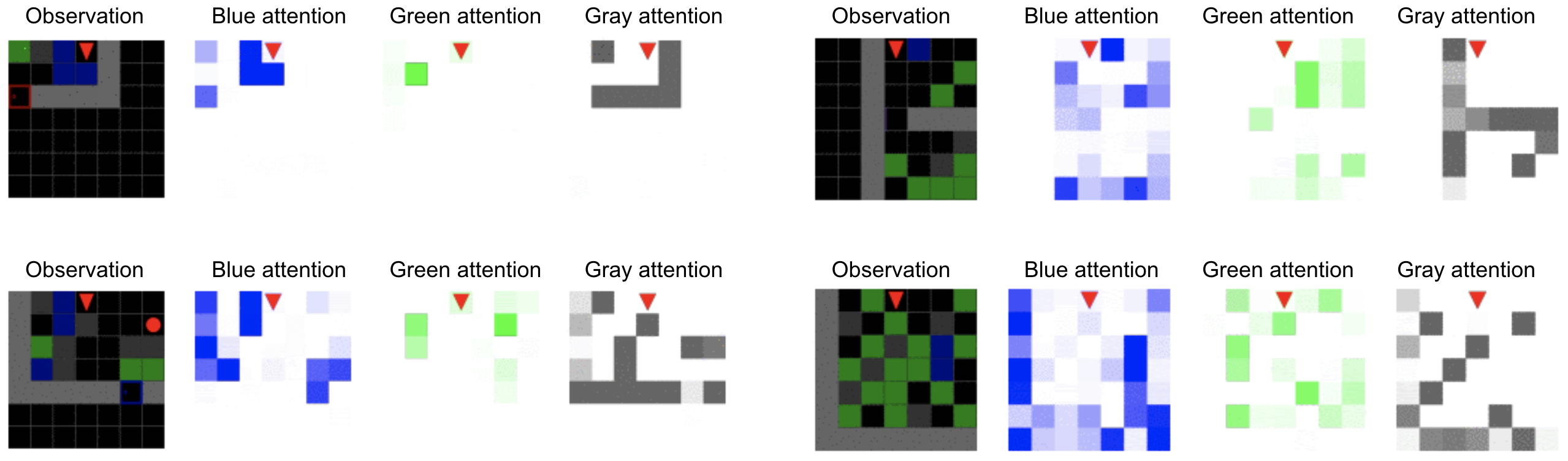} \\ \vspace{3mm}
    \includegraphics[width=\linewidth]{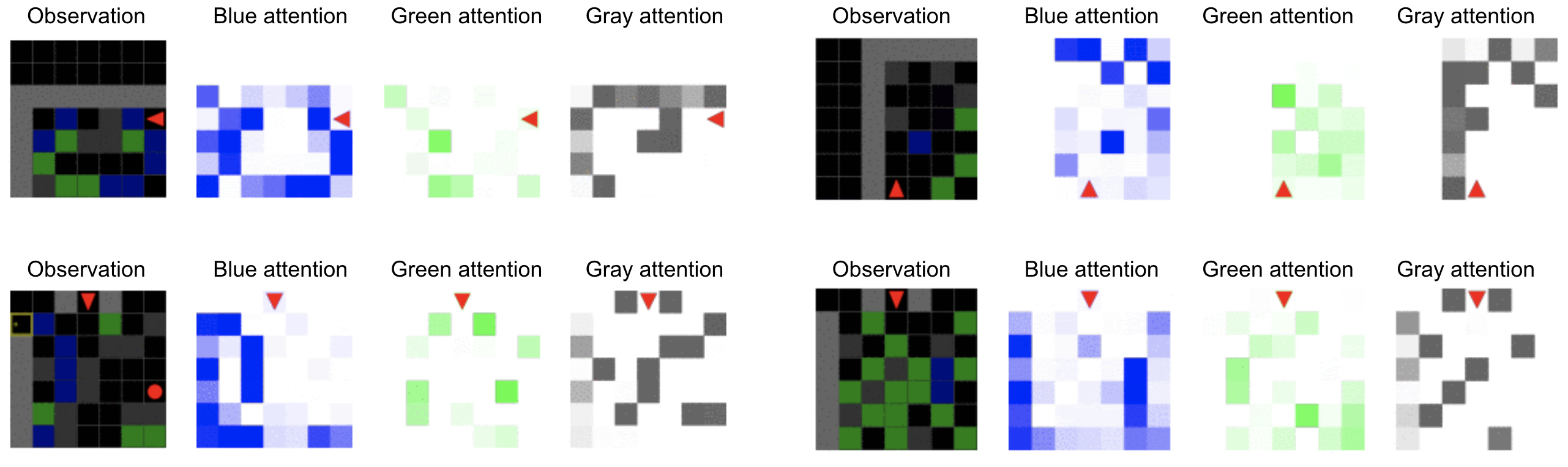}
    \caption{Visualization of attention module in learned networks trained with RL on \texttt{GotoRedBall-v2}. The first panel shows the observation from the agent's view, and the rest three panels from the right shows the attention weighting to grid positions for each sentence of description. The position of the agent in each panel is highlighted in red. }
    \label{fig:attn_viz_ap}
\end{figure}

To further verify whether performance improvements of the proposed method were due to language grounding, we provide supplementary visualizations examples for the attention weights of each sentence in the descriptions to spatial locations in Figure~\ref{fig:attn_viz_ap}.

\end{document}